\documentclass[runningheads]{llncs}
\usepackage{graphicx}
\usepackage{amsmath,amssymb} 
\usepackage{color}
\usepackage[colorlinks=true, allcolors=blue]{hyperref}
\usepackage[ruled,vlined,algo2e]{algorithm2e}
\usepackage{amsfonts}
\usepackage{multirow}
\usepackage{booktabs}
\usepackage{color,soul}
\usepackage{capt-of,etoolbox}
\usepackage{etoolbox,lipsum}
\usepackage[symbol]{footmisc}

\renewcommand{\thefootnote}{\fnsymbol{footnote}}

\newcommand\blfootnote[1]{%
  \begingroup
  \renewcommand\thefootnote{}\footnote{#1}%
  \addtocounter{footnote}{-1}%
  \endgroup
}

\begin{document}

\title{Generation of Virtual Dual Energy Images from Standard Single-Shot Radiographs using Multi-scale and Conditional Adversarial Network}

\titlerunning{Generation of Virtual Dual Energy Images}


\author{Bo Zhou\inst{1}\textsuperscript{\dag} \and
Xunyu Lin\inst{1}\textsuperscript{\dag} \and
Brendan Eck\inst{2} \and
Jun Hou\inst{2} \and
David L. Wilson\inst{2}}

\institute{Robotics Institute, School of Computer Science, Carnegie Mellon University \and
Department of Biomedical Engineering, Case Western Reserve University
\email{zhoubocp9@gmail.com} \\}

\authorrunning{B. Zhou, etc.}

\maketitle
\begin{abstract}
Dual-energy (DE) chest radiographs provide greater diagnostic information than standard radiographs by separating the image into bone and soft tissue, revealing suspicious lesions which may otherwise be obstructed from view. However, acquisition of DE images requires two physical scans, necessitating specialized hardware and processing, and images are prone to motion artifact. Generation of virtual DE images from standard, single-shot chest radiographs would expand the diagnostic value of standard radiographs without changing the acquisition procedure. We present a Multi-scale Conditional Adversarial Network (MCA-Net) which produces high-resolution virtual DE bone images from standard, single-shot chest radiographs. Our proposed MCA-Net is trained using the adversarial network so that it learns sharp details for the production of high-quality bone images. Then, the virtual DE soft tissue image is generated by processing the standard radiograph with the virtual bone image using a cross projection transformation. Experimental results from 210 patient DE chest radiographs demonstrated that the algorithm can produce high-quality virtual DE chest radiographs. Important structures were preserved, such as coronary calcium in bone images and lung lesions in soft tissue images. The average structure similarity index and the peak signal to noise ratio of the produced bone images in testing data were 96.4 and 41.5, which are significantly better than results from previous methods. Furthermore, our clinical evaluation results performed on the publicly available dataset indicates the clinical values of our algorithms.  Thus, our algorithm can produce high-quality DE images that are potentially useful for radiologists, computer-aided diagnostics, and other diagnostic tasks. Our code is available at \href{https://github.com/bbbbbbzhou/Virtual-Dual-Energy}{https://github.com/bbbbbbzhou/Virtual-Dual-Energy}
\blfootnote{\textsuperscript{\textbf{\dag}} Both authors equally contributed to this work}
\end{abstract}

\section{Introduction}
Chest radiography (CR) is the most ordered clinical imaging procedure for the initial detection of abnormality in the chest. As a noninvasive, low radiation dose, and low cost imaging modality, CR is commonly used for detecting lung disease, such as lung cancer, and diagnosing conditions such as tuberculosis and pneumonia. However, some lung lesions are extremely difficult to detect due to overlapping bone structures such as ribs and clavicles. Dual energy subtraction (DES) can separate high-density material, such as bone, from soft tissue and thus improve detection \cite{kelcz1994conventional,li2011improved}. Recent studies have also shown that DES can detect and assess coronary disease by visualizing coronary calcifications in the DE bone image \cite{zhou2016detection,zhou2018visualization,wen2018enhanced} and computer-aided disease detection tasks \cite{chen2013computerized}. DE chest radiography requires two x-ray exposures at two different tube voltages to capture two radiographs which are then linearly combined to generate bone images and soft-tissue images \cite{vock2009dual}. However, specialized equipment is required for DES and radiation dose is nearly doubled as compared to traditional CR. In addition, the organ motion artifacts caused by heartbeat and lung breathing motion during the time-lapse between the two kVp x-ray exposures can contaminate the image quality of DE radiographs. The ability to generate bone and soft tissue images from standard, single-shot CR would benefit this diagnostic imaging procedure.

In this work, our goal is to develop a deep learning algorithm for generating virtual DE images from a standard, single-shot chest radiograph. We collected a large number of real two-exposure DE chest radiographs as training data for this task. The Convolutional Neural Network (CNN) has achieved significant success in medical image analysis field for tasks, such as chest disease detection/classification \cite{wang2017chestx}; CT pulmonary nodule detection \cite{shin2016deep}; automatic organ segmentations \cite{roth2015deeporgan}, etc. However, predicting DE images using deep models remains a challenge. The structure and contextual information of a large receptive field in standard radiograph should be extracted by CNN model in a maximal manner to determine whether bony component are present and to predict the corresponding distribution. If the CNN model for fine-scale prediction is in a fully convolutional form, the size of CNN model would become very large with excessive number of parameters to learn, making it difficult to train.

\begin{figure*}[!htb]
\centering
\includegraphics[width=0.98\textwidth]{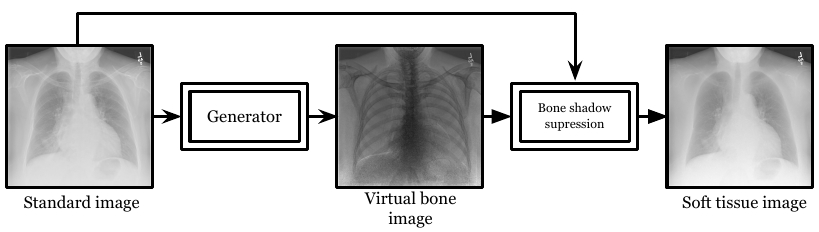}\\
\caption{Overall processing pipeline. A virtual DE bone image is generated from the MCA-Net generator. Given the bone image, bone in the standard/high kVp image is suppressed to produce a virtual soft tissue image.}
\label{fig:pipeline}
\end{figure*}

In order to avoid training a very large CNN model and  efficiently train a CNN model to predict fine-scale DE information, we propose a deep model to generate high-quality virtual DE images from a standard, single-shot chest radiograph. Our model is based on a multi-scale and conditional adversarial network. The general pipeline of the algorithm is shown in Figure \ref{fig:pipeline}. The algorithm is comprised of two parts: 1) the bone image generator using the multi-scale fully convolutional network, and 2) the soft tissue image generator which applies bone suppression on standard images. We introduce the concept of conditional adversarial loss in training the bone image generator \cite{isola2017image} so that high frequency information and details are learned and preserved in the virtual bone image. To produce the virtual soft tissue image, an adapted edge and shadow suppression algorithm using cross projection tensor \cite{agrawal2006edge} is applied to suppress bone in the standard radiograph. Outputs from the algorithm are compared to a set of test data comprised of DE patient images. The algorithm performance is also evaluated by comparing to other algorithms that are applicable for virtual DE image generation.

\subsection{Related Work}
Bone suppression techniques have been developed to improve the diagnostic quality of CR exams for interpretation of disease. Current methods can be summarized into two types: learning-based methods and statistical analysis based methods. For learning-based methods, one of the early works was MTANN, proposed in \cite{suzuki2004suppression,suzuki2006image}, which uses traditional fully-connected neural networks with one hidden layer to predict the bone signal from a standard image. Then, the predicted bone signal can be subtract from standard image to generate an image similar to DE soft tissue image. Similar to one of the ideas in our network structure, MTANN was trained under a multi-resolution style. Each MTANN was trained separately for a certain resolution with corresponding re-sampled image using multi-resolution decomposition, and the prediction targets were the intensity values of single pixels from the DE bone images. At the testing stage, the trained MTANNs generate multi-resolution bone images and these images are combined together to produce a high-resolution bone image. Later, \cite{chen2014separation} further improved the MTANN to separate bone from soft tissue by training MTANN in different anatomical regions and producing DE images using total variation optimization. Such methods were extended to be used in portable CR systems \cite{chen2016enhancement}. Another one of the early works was the k-nearest neighbors (kNN) regression with optimized local feature \cite{loog2006filter} which used a linear dimensionality reduction method for local image features to optimize the performance of kNN regression for the prediction of bone images. However, this method cannot completely suppress the rib signal and requires a relatively long computation time for kNN regression. Our proposed deep model aims to address the limitations of previous work by generating high-resolution virtual DE images with relatively short computation times.

Besides from the learning-based methods, several statistical analysis based methods for suppressing bone structures without the supervision from training data have been studied. \cite{simko2009elimination} proposed a clavicle suppression algorithm which works by first generating a bone image from a gradient map modified along the bone border direction and then creating a soft-tissue image by subtraction of the bone image from the standard image. Later, \cite{hogeweg2013suppression,rasheed2007rib} proposed blind-source signal separation algorithms for suppression of bone structures in standard chest radiography. \cite{zhou2018visualization} presented a ribcage segmentation algorithm based on Active Appearance Model that can accurately estimate the rib border and suppress the bone signal from standard radiography based on this prediction. In general, these statistical based methods require accurate segmentation and border annotations for the target structures, which is challenging to acquire. Although all have shown improvement in thoracic disease detection, many image details are suppressed and these methods require substantial hyper-parameter tuning.

In this case, deep learning has been successfully applied in image classification, image segmentation, and image-to-image translation \cite{lecun2015deep}. Deep learning also yields similar improvements in performance in the medical vision field for anatomical and pathological structures detection and segmentation tasks.\cite{wang2017chestx,shin2016deep,roth2015deeporgan,mirza2014conditional} The boost of hardware and algorithms for deep learning have given the possibility of training CNN with many layers on large-scale datasets. Our work is closely related to the current state-of-the-art CNN models for image synthesis \cite{isola2017image} and image transformation \cite{ledig2016photo}. \cite{isola2017image} proposed a conditional GAN (cGAN) that can generate realistic images given the outlines of the target image as the prior condition. \cite{ledig2016photo} employed a GAN structure for generating high-resolution images from nominal resolution images. Our proposed method for generation of virtual DE images benefits from these excellent works on the application of GAN. The standard radiographs were used as a strong prior condition for generating virtual DE images using structure similar to cGAN \cite{isola2017image}. The adversarial network structure also helps generation of super-resolution DE images (normally around 2022 pixels$\times$ 2022 pixels). In this paper, a customized GAN structure was designed for the generation of virtual DE bone images.

\subsection{Contributions}
In summary, our contributions are listed as follows:
\begin{itemize}
\item We propose a novel algorithm for generation of virtual dual energy images from single-shot standard radiographs based on a customized adversarial network. 
\item We collected a relatively large number of clinical images and demonstrated our deep model's superior performance compared to current approaches.
\item We evaluated our proposed algorithm on a public chest radiograph dataset and obtained clinical values on diagnostic task of our algorithm.
\end{itemize}

\section{Methods}
\begin{figure*}[!htb]
\centering
\includegraphics[width=1\textwidth]{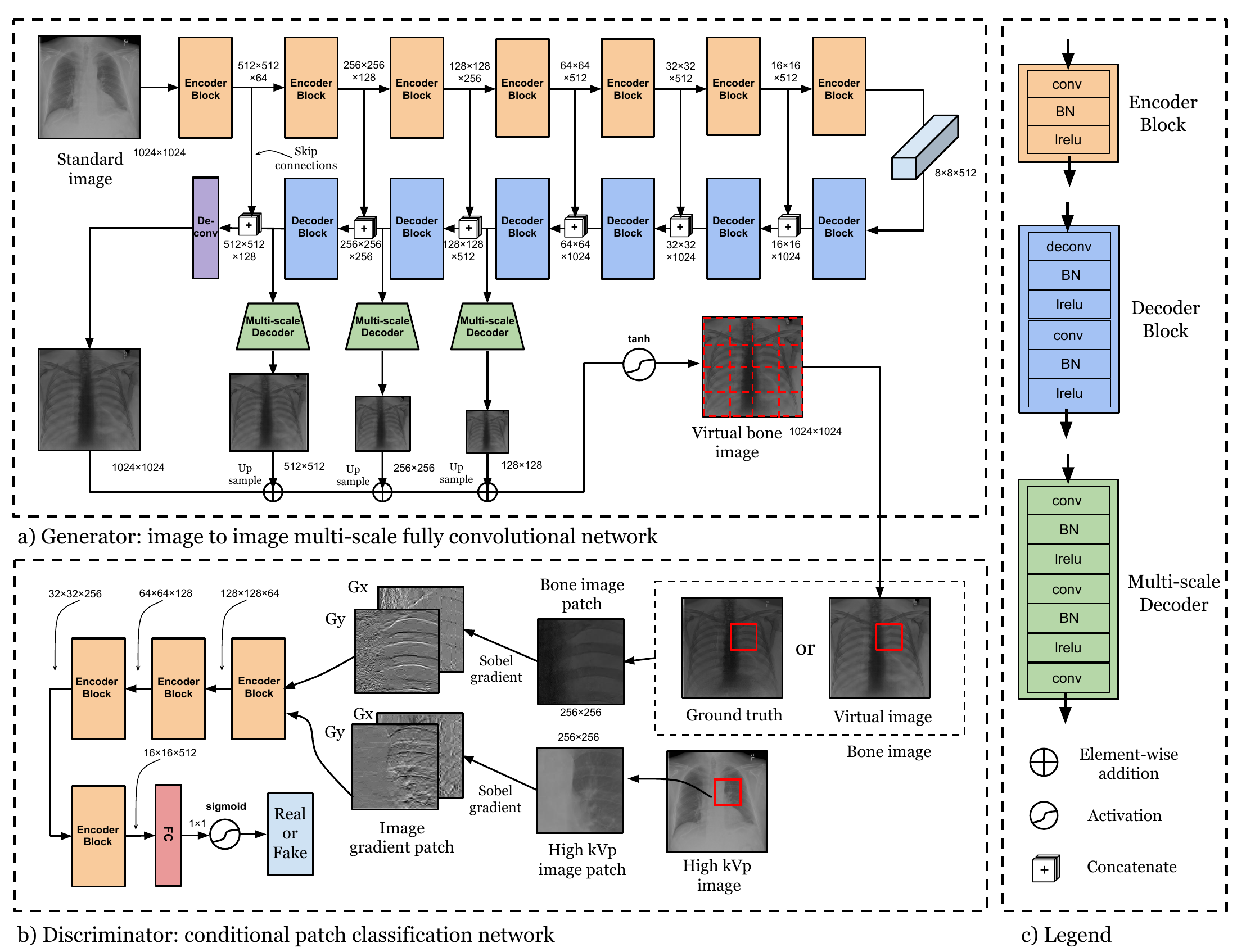}\\
\caption{Illustration of the MCA-Net architecture: a) the multi-scale generator computes a virtual bone image from a standard/high kVp image; b) the conditional patch discriminator distinguishes real and generated images by looking at the bone image gradient patch and conditioning on the standard image gradient patch; c) legend of operations.}
\label{fig:architecture}
\end{figure*}

We propose a novel algorithm to generate virtual bone and soft tissue images from a standard, single-shot CR. It consists of two parts: 1) generation of a virtual bone image from a standard CR using Multi-scale and Conditional Adversarial network (MCA-Net); 2)  generation of a virtual soft tissue image using bone suppression of the standard CR with the virtual bone image. In order to generate bone images with both well-shaped general appearance and fine-grained details, we introduced MCA-Net with multi-scale generations and patch-wise adversarial learning \cite{isola2017image}. The architecture of MCA-Net is shown in Figure \ref{fig:architecture}, with details described in the following sections.

\subsection{\label{Network}Multi-scale and Conditional Adversarial Network (MCA-Net)}

\subsubsection{Multi-scale generator}
The multi-scale architecture is motivated by the observation that a coarse-to-fine generation can be beneficial for generating high resolution images \cite{karras2017progressive}. Therefore, the multi-scale generator is designed to first generate a low resolution image capturing coarse appearances, then to add finer details generated from a higher resolution image of higher network level outputs. The final generation utilizes all of the images at different resolutions.

As shown in \ref{fig:architecture}a), the multi-scale generator follows the encoder-decoder bottleneck architecture. The encoder is formed by a set of convolution operations to map standard images to deep features. The decoder incorporates both convolution and deconvolution operations to generate bone images at different scales from these features. Images at different scales are added together element-wise and followed by a $\tanh$ activation function to form the final output. Skip connections \cite{ronneberger2015u} between encoder-decoder and multi-scale supervision in the network provide more feedback to aid back-propagation and reduce the effect of gradient vanishing in the shallow layers.

\subsubsection{Adversarial learning and conditional patch-discriminator}
Adversarial learning \cite{goodfellow2014generative} is a game-theory-based learning scheme consisting of two models, a discriminator and a generator, which are trained to combat each other. As shown in Figure \ref{fig:architecture}b), the discriminator tries to distinguish ground truth bone images from virtual bone images (output by the generator) by conditioning on the standard images. Adopted from \cite{isola2017image}, the discriminator only takes image patches as input which gives two benefits: 1) less memory usage, and 2) improving network capability to identify finer details in one image patch.

Instead of feeding raw image patches, we feed the discriminator with image gradients computed with a Sobel filter in both x and y directions. Using an image gradient can enhance high frequency signals and result in sharper generated images. It has been shown that adversarial learning can help to generate high frequency signals that are usually not captured by $L_1$ loss, such as sharp edges and fine textures \cite{isola2017image}.

\subsubsection{Hybrid loss function} With only adversarial learning, artifacts may be present in the generated DE bone image \cite{isola2017image}. To alleviate this, we combine $L_1$ loss together with adversarial loss to form a hybrid loss function, where $L_1$ and adversarial loss contribute to the low and high frequency information to the image generation, respectively. Specifically, $L_1$ loss helps in learning the general appearance of the skeleton while adversarial loss emphasizes sharp edges and aims to preserve subtle calcium signals. We denote $L_1$ loss as $L_{l_1}$ and adversarial loss as $L_{adv}$, calculated according to:
\begin{equation}
\label{equa_adv}
L_{adv}(G,D)=-\mathbb{E}_{I_H,I_B}\left[\log(D(I_H,I_B))\right]-\mathbb{E}_{I_H}\left[\log(1-D(I_H,G(I_H)))\right]
\end{equation}

\begin{equation}
\label{equa_l1}
L_{l_1}(G)=\mathbb{E}_{I_H,I_B}\left[\left\Vert I_B-G(I_H)\right\Vert_1\right]
\end{equation}

\noindent
where $D$ and $G$ denote the discriminator and generator, respectively. $G$ is a function of the standard, high x-ray tube voltage chest radiograph, $I_H$, and generates a virtual bone image for comparison by the discriminator. $D$ conditions on $I_H$ and takes either the ground truth bone image, $I_B$, or virtual bone image generated by $G$ as input and determines whether the input is a real bone image or virtual bone image. A value of $1$ indicates a real image and 0 indicates a virtual image. Overall, we seek G minimizing the hybrid loss of equation (\ref{equa_adv}) and (\ref{equa_l1}) while D is maximizing equation (\ref{equa_adv}). The final objective of our system is to obtain the optimal generator $G^*$ that satisfies:
\begin{equation}
G^*=arg\min_{G}(\max_{D}L_{adv}(G,D)+\lambda L_{l_1}(G))
\end{equation}

\noindent
where $\lambda$ controls the importance given to $L_1$ loss. As discussed above, the discriminator only processes patches instead of the entire image. Thus, the inputs to $D$ are three patches: one from $I_H$, one from $I_B$, and one from the generated virtual bone image, $G(I_H)$.

\subsection{\label{CrossTensor}Bone suppression with Cross Projection Tensor}
\begin{figure*}[!htb]
\centering
\includegraphics[width=1\textwidth]{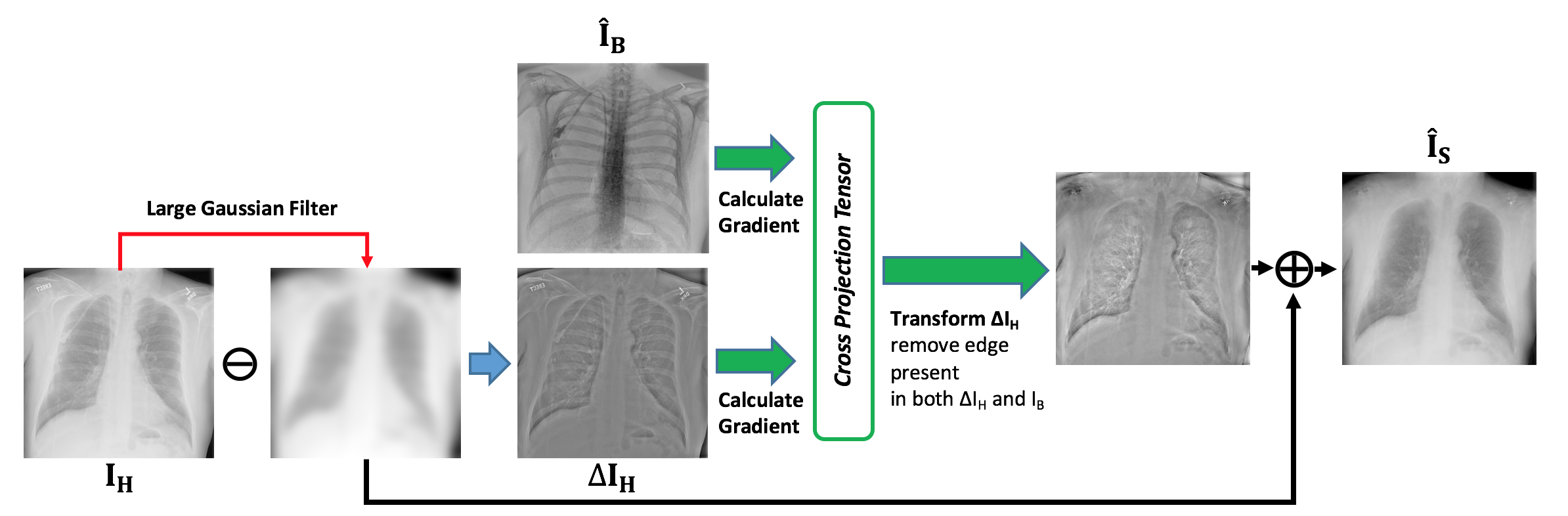}\\
\caption{Image processing pipeline of bone suppression with cross projection tensor.}
\label{fig:crosstensor}
\end{figure*}

Given the virtual bone image $\hat{I}_{B}$ and standard image $I_{H}$, we generate a virtual soft tissue image $\hat{I}_{S}$. Bone signals from $\hat{I}_{B}$ are treated as shadows in $I_{H}$ and are removed by our processing pipeline as shown in Figure \ref{fig:crosstensor}. Bone shadows are removed using the cross projection tensor algorithm proposed in \cite{agrawal2006edge}. First, we estimate the general soft tissue profile by convolving a $201 \times 201$ Gaussian filter with $\sigma=50$, to $I_{H}$. Then, the Gaussian filtered $I_{H}$ is subtracted from $I_{H}$ to yield $\Delta I_{H}$ which only contains high frequency signal from soft tissue and bone. After that, the cross projection tensors are computed with the gradient images of $\hat{I}_{B}$ and  $\Delta I_{H}$.  The gradient field of $\Delta I_{H}$ is transformed by the cross projection tensor, removing edges from $\Delta I_{H}$ that are also present in $\hat{I}_{B}$. Finally, we add the general soft tissue profile with the high frequency soft tissue to produce the virtual soft tissue image $\hat{I}_{S}$. 

\section{Data and Training Details}
We collected 210 posterior-anterior DE chest radiographs with a two-shot DE digital radiography system (Discovery XR656, GE Healthcare). The data was acquired using a 60 kVp exposure followed by 120 kVp exposure procedure with 100 ms between exposures. The sizes of the images ranged from $1300 \times 1400$ to $2022 \times 2022$ pixels (approximately 39.4 cm by 39.4 cm). We split the dataset into a training set of 170 cases and a testing set of 40 cases. More information about our DE dataset is summarized in Table \ref{tab:dataset}. Data augmentation was performed on the training set by randomly 1) translating the images in x,y directions from [-80, 80] pixels; and 2) rotating the images from [-15, 15] degrees about the image center. A total of 3906 cases were augmented to train the network, and evaluations were made on the testing set.

\begin{table} [!htb]
\centering
\caption{Statistical information of our DE datasets}
\label{tab:dataset}
\setlength{\tabcolsep}{7pt}

\begin{tabular}{l c c c c c c c c}
\hline
\rule{0pt}{1.1\normalbaselineskip}
Dataset & Total & \multicolumn{4}{c}{Age} & \multicolumn{2}{c}{Gender} \\ \cmidrule(r){3-6} \cmidrule(r){7-8}
        &       & $\leqslant20$ & $20\sim40$ & $40\sim60$ & $\geqslant60$ & Male & Female \\ [0.1cm]
\hline
\rule{0pt}{1.1\normalbaselineskip}
Training      & $170$ & $5$ & $36$ & $78$ & $51$ & $114$ & $56$ \\
\rule{0pt}{1.1\normalbaselineskip}
Test       & $40$ & $1$ & $9$ & $17$ & $13$ & $32$ & $8$ \\ [0.1cm]
\hline

\end{tabular}
\end{table}

The training scheme of our adversarial network is depicted in Algorithm \ref{alg:train_alg}. 
Before training, we normalized all images to the range of $[-1,1]$ per image by Equation (\ref{equ:normalize}), where $I_{raw}$ and $I_{norm}$ are unnormalized and normalized images, respectively. Our network directly takes normalized standard images to generate normalized virtual bone images. Ground truth intensity maximum and minimum is then applied to the normalized virtual bone image to recover the final result.

\begin{equation}
I_{norm}^{x,y}=2\times\dfrac{I_{raw}^{x,y}-\min_{\forall i,j} I_{raw}^{i,j}}{\max_{\forall i,j} I_{raw}^{i,j}}-1
\label{equ:normalize}
\end{equation}

We observed synthesis artifacts in virtual images when $\lambda$ was too small. We found that $\lambda=1000$ provided a good balance between low and high frequency information in generations. $N_G$ and $N_D$ were chosen to balance the ability of two networks. Training the generator with more iterations than the discriminator, $N_G>N_D$, gave us better results (including sharper images and fewer artifacts). Although optimal values for $N_D$ and $N_G$ could be further optimized, we observed that a ratio $\frac{N_G}{N_D}$ around 3 generally gave good quality generations. The network was trained with batch size of 3 and learning rate $10^{-4}$ for 100 epochs.

\begin{algorithm2e}[!htb]
\caption{Training MCA-Net for one epoch}\label{alg:train_alg} 
\KwIn 
{%
Weights of G: $W_G$; weights of D: $W_D$; i-th mini-batch of standard images and bone images: $I_H^i$, $I_B^i$ ($i=1,\ldots,n$) 
}
\textbf{Initialize:} $i=1$; \# of iterations to train G: $N_G=3$; \# of iterations to train D: $N_D=1$; hybrid loss weight: $\lambda=1000$\;
\While{$i \leq n$}
{
      \For{$k = 1$ to $N_D$}
      {
        $L_D \gets -\mathbb{E}_{I_H^i,I_B^i}\left[\log(D(I_H^i,I_B^i))\right]-\mathbb{E}_{I_H^i}\left[\log(1-D(I_H^i,G(I_H^i)))\right]$\;
        $W_D \gets Optimizer(L_D, W_D)$\;
        $i \gets i + 1$\;
      }
      \For{$l = 1$ to $N_G$}
      {
        $L_G \gets -\mathbb{E}_{I_H^i}\left[\log(D(I_H^i,G(I_H^i)))\right] + \lambda\mathbb{E}_{I_H^i,I_B^i}\left[\left\Vert I_B^i-G(I_H^i)\right\Vert_1\right] $\;
        $W_G \gets Optimizer(L_G, W_G)$\;
        $i \gets i + 1$\;
      }
}
\end{algorithm2e} 

\section{Experiments and Results}
\subsection{Algorithm Evaluation}
\begin{figure*}[!htb]
\centering
\includegraphics[width=1\textwidth]{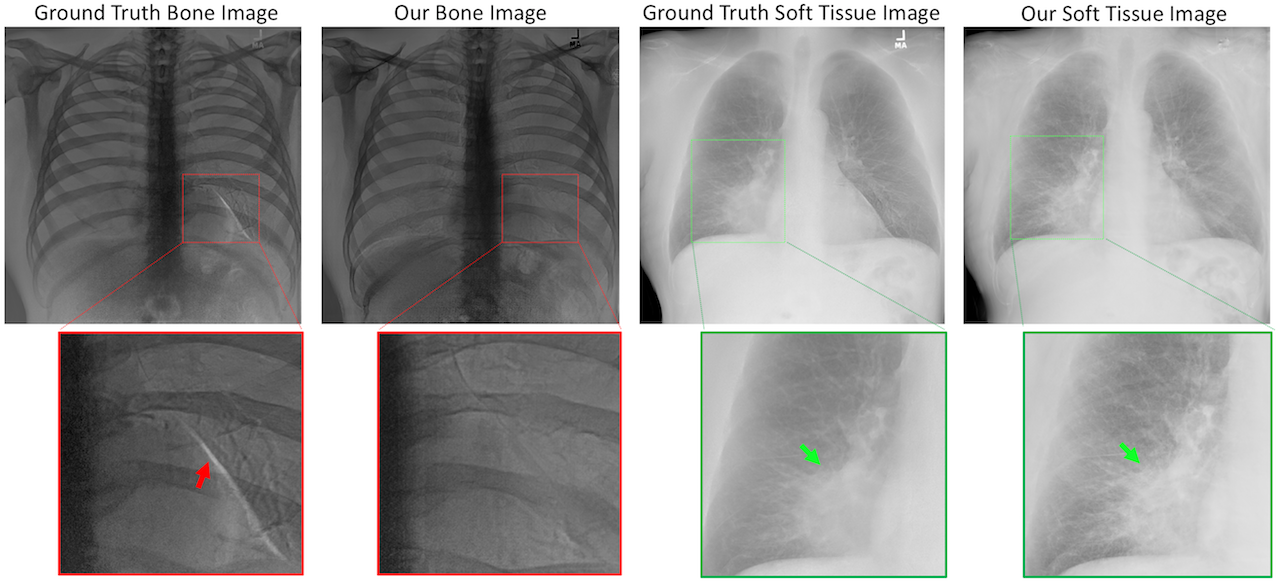}\\
\caption{Dual energy bone and soft tissue images of a 64-year-old male patient with cardiogenic pulmonary edema (CPE: green arrows) generated from DES and the virtual DES algorithm. Motion artifact (red arrow) in DES, due to cardiac motion between exposures, is significantly suppressed using from the virtual DES algorithm. CPE in the virtual soft tissue image has sharper visualization of pulmonary veins compared to ground truth.}
\label{fig:ourimages}
\end{figure*}

Figure \ref{fig:ourimages} shows a testing data example of virtual DE bone and soft tissue images from MCA-net compared to ground truth DE images. Cardiac motion artifacts are significantly reduced in the virtual DE images. High-quality and high-resolution virtual bone and soft tissue images with subtle details are produced. 

\begin{figure*}[!htb]
\centering
\includegraphics[width=1\textwidth]{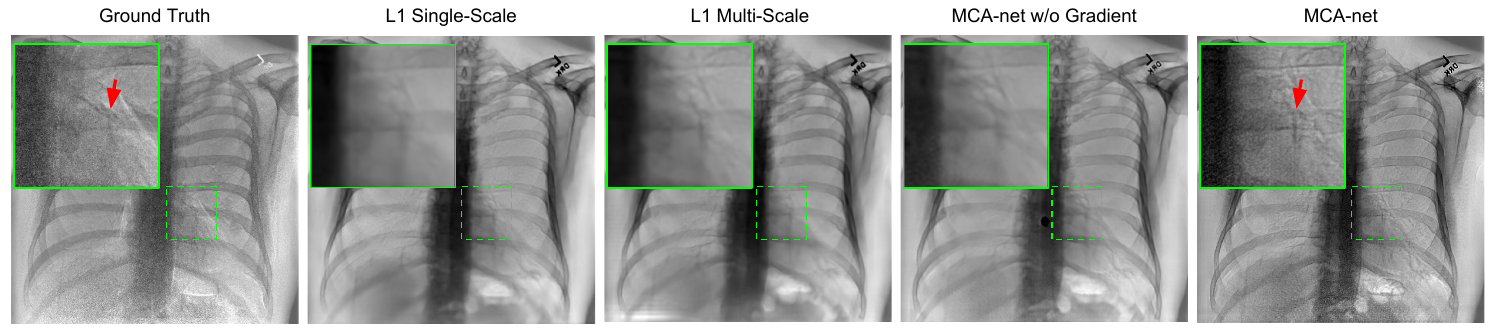}\\
\caption{Comparison of DE bone images generated with or without multi-scale or adversarial training. MCA-Net can preserve subtle calcium signal such as coronary calcium (red arrow).}
\label{fig:bone_example}
\end{figure*}

\begin{table} [!htb]
\centering
\caption{Comparison of algorithm performance with different training settings}
\label{tab:t1}
\setlength{\tabcolsep}{7pt}

\begin{tabular}{l c c c c c}
\hline
\rule{0pt}{1.1\normalbaselineskip}
Network Structure             & \multicolumn{2}{c}{PSNR (dB)} & \multicolumn{2}{c}{SSIM} \\ \cmidrule(r){2-3} \cmidrule(r){4-5}
                              & $I_{Bone}$ & $I_{Soft}$ & $I_{Bone}$ & $I_{Soft}$  \\ [0.1cm]
\hline
\rule{0pt}{1.1\normalbaselineskip}
$l_{1}$ loss + single scale      & $28.2 \pm 4.5$ & $22.2 \pm 3.4$ & $86.4 \pm 3.1$ & $80.3 \pm 3.4$ \\
\rule{0pt}{1.1\normalbaselineskip}
$l_{1}$ loss + multi-scale       & $34.9 \pm 3.6$ & $29.2 \pm 4.1$ & $88.3 \pm 2.8$ & $81.4 \pm 3.7$ \\
\rule{0pt}{1.1\normalbaselineskip}
MCA-Net w/o gradients        & $35.2 \pm 5.1$ & $29.8 \pm 3.8$ & $88.5 \pm 2.5$ & $81.2 \pm 4.3$ \\
\rule{0pt}{1.1\normalbaselineskip}
\textbf{MCA-Net w/ gradients}  & $\bold{41.5 \pm 2.1}$ & $\bold{39.7 \pm 1.8}$ & $\bold{93.4 \pm 1.4}$ & $\bold{88.4 \pm 3.4}$ \\ [0.1cm]
\hline

\end{tabular}
\end{table}

Two evaluation metrics were used to assess the performance of our algorithm. Given a virtual image $\hat{I}$ and ground truth image $I$, we can calculate the Peak Signal-to-Noise Ratio (PSNR) and Structure Similarity index (SSIM). PSNR is commonly used for evaluating quality of image compression/reconstruction, where our algorithm can be viewed as a compression/reconstruction process. SSIM is used for measuring the structure similarity between two images. We evaluated the improvement in virtual image generation from the conditional adversarial loss and multi-scale generator architectures. We compared the PSNR and SSIM values of images generated with and without those network components. The quantitative results are listed in Table \ref{tab:t1} using the testing dataset. The average PSNR values of virtual bone and soft tissue images produced by MCA-net are 41.5 and 39.7. The average SSIM values of bone and soft tissue images generated by MCA-net are 93.4 and 88.4. Adding the conditional adversarial loss and the multi-scale output in generator, the PSNR and SSIM were significantly improved. A visual comparison of images generated with and without those network components is shown in Figure \ref{fig:bone_example}. Important calcium signals, such as coronary artery calcification, are preserved in the virtual bone image which are potentially useful for evaluating cardiovascular diseases.

We compared the prediction performance of the models by comparing to other methods (Table \ref{tab:t_compare}) using the same data. In addition to SSIM and PSNR, the Relative Mean-Absolute-Error (RMAE) is used as an additional metric. Please note MTANN \cite{suzuki2006image} and Vis-CAC \cite{zhou2018visualization} don't generate virtual DE bone image, so there is no comparison available for virtual DE bone images. The results show that our algorithm has better performance in the generation of virtual DE images and is able to produce high-quality DE bone images.




\begin{table} [!htb]
\centering
\caption{Comparison of algorithm performance with previous works}
\label{tab:t_compare}
\setlength{\tabcolsep}{5.3pt}

\begin{tabular}{l c c c c c c c}
\hline
\rule{0pt}{1.1\normalbaselineskip}
Methods                       & \multicolumn{2}{c}{RMAE} & \multicolumn{2}{c}{SSIM} & \multicolumn{2}{c}{PSNR (dB)} \\ \cmidrule(r){2-3} \cmidrule(r){4-5} \cmidrule(r){6-7}
                              & $I_{Bone}$ & $I_{Soft}$ & $I_{Bone}$ & $I_{Soft}$ & $I_{Bone}$ & $I_{Soft}$ \\ [0.1cm]
\hline
\rule{0pt}{1.1\normalbaselineskip}
MTANN [2]        & \scriptsize N/A & \scriptsize $28.3 \pm 8.4$ & \scriptsize N/A & \scriptsize $84.3 \pm 4.7$ & \scriptsize N/A & \scriptsize $39.2 \pm 3.1$ \\
\rule{0pt}{1.1\normalbaselineskip}
Vis-CAC [3]     & \scriptsize N/A & \scriptsize $16.8 \pm 7.1$ & \scriptsize N/A & \scriptsize $80.3 \pm 3.4$ & \scriptsize N/A & \scriptsize $38.8 \pm 4.2$ \\
\rule{0pt}{1.1\normalbaselineskip}
Ours                                    & \scriptsize $\bold{8.3 \pm 2.3}$ & \scriptsize $\bold{12.1 \pm 4.3}$ & \scriptsize $\bold{93.4 \pm 1.4}$ & \scriptsize $\bold{88.4 \pm 3.4}$ & \scriptsize $\bold{41.5 \pm 2.1}$ & \scriptsize $\bold{39.7 \pm 1.8}$ \\ [0.1cm]
\hline

\end{tabular}
\end{table}

\begin{figure*}[!htb]
\centering
\includegraphics[width=1\textwidth]{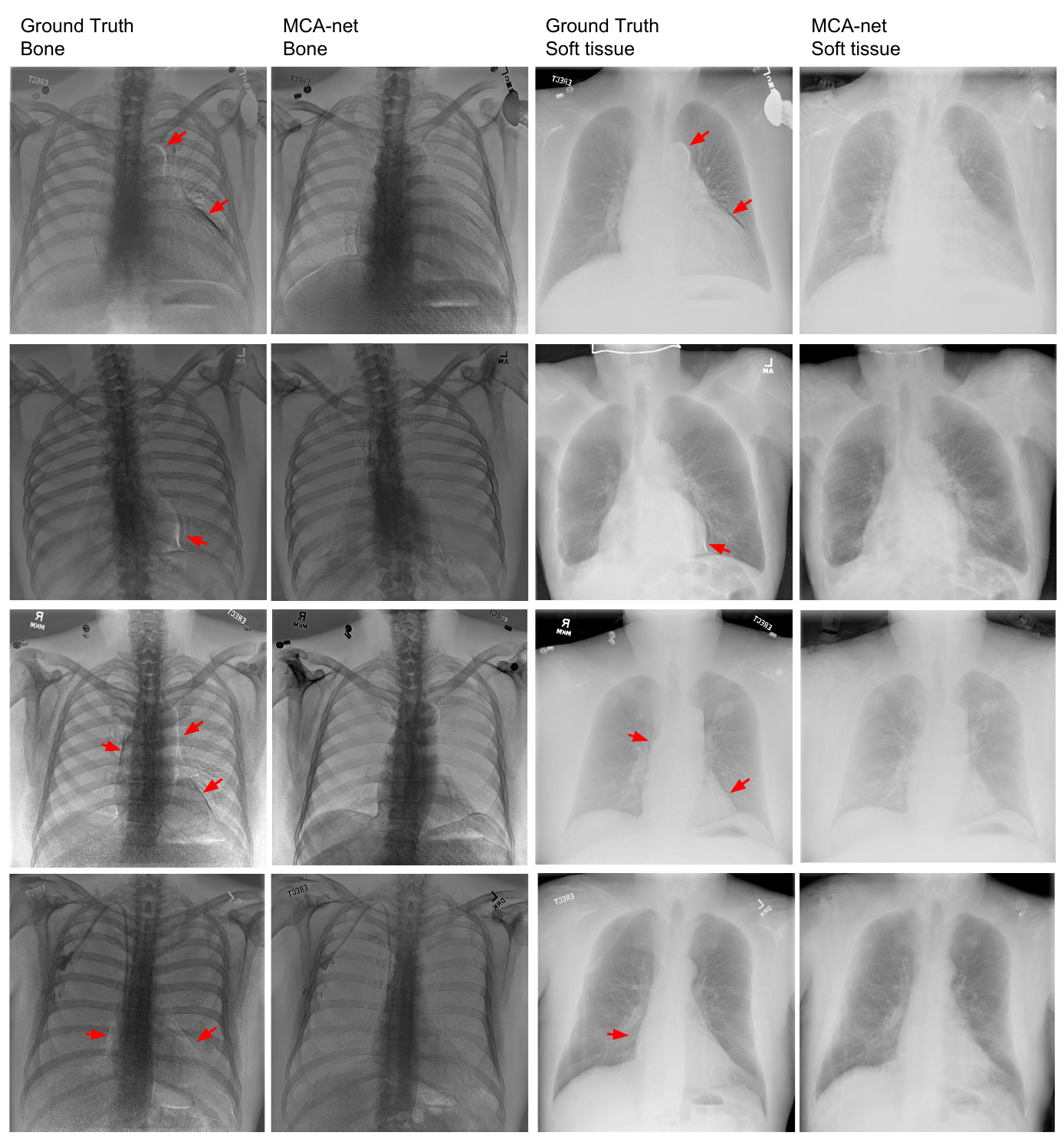}\\
\caption{Additional examples of comparison results between ground truth and MCA-net's results. The cardiac motion artifact (red arrows) along the heart boundary on both DE bone image and DE soft tissue images are significantly reduced in our virtual images. Mitigation of motion artifacts led to improvements in overall image quality.}
\label{fig:sm}
\end{figure*}

We included additional examples of comparison results between ground truth DE images and virtual DE images as shown in Figure \ref{fig:sm}. Overall image characteristics are captured by the algorithm, such as dileneation of the spine, ribs, and clavicles in the virtual bone images, and visualization of pulmonary vessels in the soft tissue images. Cardiac motion artifact was present in all cases and is significantly suppressed in the generated virtual dual energy images.

\subsection{Clinical Evaluation}
To evaluate the clinical application of our algorithm, we used the publicly available Japanese Society of Radiological Technology (JSRT) database \cite{schilham2006computer}. The Posterior-Anterior standard chest radiographs in the database were collected from 14 medical institutions by use of screen-film systems over a period of three years. All nodules present in the radiographs were confirmed by CT, and the locations of the nodules were confirmed by chest radiologists. The images were digitized to yield 12-bit images with a resolution of $2048 \times 2048$ pixels. The size of a pixel was $0.175 \times 0.175$ mm. This database contains 93 normal cases and 154 cases with confirmed lung nodules. 

\begin{figure*}[!htb]
\centering
\includegraphics[width=1\textwidth]{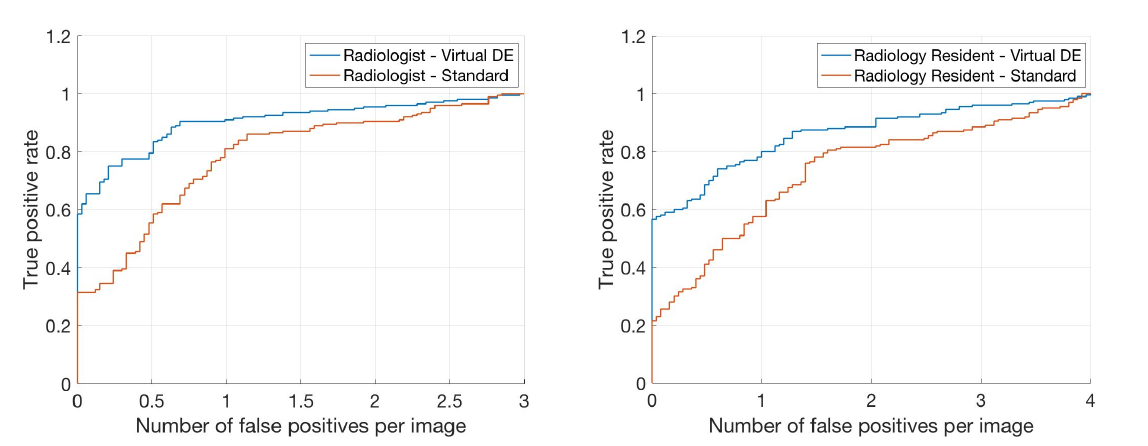}\\
\caption{Comparison of FROC curves on lung nodule localization using standard radiography and virtual DE radiography. Left: the chest radiologist results; Right: the radiology resident results. Use of the virtual DE images improved detection and localization of lung nodules.}
\label{fig:froc}
\end{figure*}

We recruited one chest radiologist and one experienced radiology resident to evaluate our algorithm's clinical application of lung nodule detection. The performance was evaluated by use of the Free-response Receiver Operating Characteristic (FROC) analysis \cite{chakraborty2007spatial}. The radiologists were asked to classify and score the data into lesion and non-lesion localizations. The scoring was done by choosing an acceptance-radius and classifying marks within the acceptance-radius of lesion centers as lesion localizations, and all other marks are classified as non-lesion localizations. The scored data is plotted as a FROC curve as shown in Figure \ref{fig:froc}, essentially a plot of appropriately normalized numbers of lesion localizations vs. non-lesion localizations. For both the radiologist and the radiology resident, the virtual DE images generated from our MCA-net shows significant improvement in localization of lung nodules. For the radiologist, we achieved sensitivity=0.91 using our virtual DE images that is significantly higher than the sensitivity=0.81 using standard images when FP=1. The detailed FROC analysis is shown in Table \ref{tab:t4}.

\begin{table} [!htb]
\centering
\caption{Mean bootstrap values are given for sensitivity in our FROC analysis. 95\textsuperscript{th}-percentile confidence intervals obtained through bootstrapping are shown between brackets. (FP=False positive detections per image) }
\label{tab:t4}
\setlength{\tabcolsep}{7pt}

\begin{tabular}{l c c c c c}
\hline
\rule{0pt}{1.1\normalbaselineskip}
\textbf{FROC}             & \multicolumn{2}{c}{Radiologist} & \multicolumn{2}{c}{Radiology Resident} \\ \cmidrule(r){2-3} \cmidrule(r){4-5}
\textbf{Analysis}                         & $1$ FP & $2$ FP & $1$ FP & $2$ FP  \\ [0.1cm]
\hline
\rule{0pt}{1.1\normalbaselineskip}
Standard      & $0.81(0.69\mbox{-}0.85)$ & $0.88(0.82\mbox{-}0.91)$ & $0.59(0.50\mbox{-}0.79)$ & $0.80(0.72\mbox{-}0.83)$ \\
\rule{0pt}{1.1\normalbaselineskip}
Virtual-DE        & $0.91(0.81\mbox{-}0.96)$ & $0.95(0.82\mbox{-}0.98)$ & $0.81(0.76\mbox{-}0.88)$ & $0.89(0.78\mbox{-}0.92)$ \\ [0.1cm]
\hline

\end{tabular}
\end{table}

\section{Discussion}
In this paper, we presented a deep network to generate virtual dual energy images from standard chest radiographs based on a multi-scale conditional adversarial network architecture (MCA-Net). According to the quantitative and qualitative results, we demonstrated our model can generate high-quality virtual DE images, and obtained significant clinical values. Several customized strategies may contribute to the effectiveness of our method. Firstly, the multi-scale fully connected network is used as the core unit for predicting the virtual bone image. The end-to-end training for large-scale samples allows it to extract effective image features from different resolutions and to have good prediction generalizablity. Secondly, the training of the multi-scale generator was reinforced with adversarial training from another deep network, so that it was able to learn how to preserve subtle signals, such as coronary calcifications, and other high frequency signals in the virtual bone image. Thirdly, adversarial training in the gradient domain may further improve the core CNN to learn the mapping between the standard image and the virtual DE images. 

Compared with the two-shot dual energy technique, the proposed MCA-net method generates DE images with fewer motion artifacts in the bone and soft tissue images. Given that majority of our dataset doesn't suffer from severe motion artifacts and was used as the training data, we assumed only the mapping between the regions of standard image and the corresponding bone information without motion artifacts was learned by the model. However, even the current DE techniques cannot perfectly separate the bone and soft tissue information from x-ray. Therefore, our method trained on current DE data were not perfect. In some case, some residual bone edge signal remained and can be observed in our virtual DE soft tissue images. Compared with similar previous works on generation of virtual DE soft tissue image through machine learning and image processing techniques, our method was trained with a relatively large scale of DE training data. For instance, there were much fewer DE training cases in \cite{suzuki2006image,chen2014separation}. Our deep model trained on relatively large-scale dataset can produce more reliable and robust virtual DE image, and generate both DE bone and soft tissue images. In addition, our method doesn't require segmentation of the lung field and the contrast normalization process for the input of standard image.

Besides from evaluating the prediction performance of our model, the clinical usefulness of our method was also assessed. In the task of localizing the lung nodule on a public dataset (JSRT), our results demonstrated that our virtual DE images can significantly improve the radiologist's performance on finding the lung nodule. Given our model's ability of preserving useful signal, such as tissue contrast and object structure, we expect similar improvement on other disease detection using our technique. However, the clinical examination of a comparison of the virtual DE images with standard image for these disease should be conducted with larger population of radiologists. Moreover, we believe a more reliable evaluation of the performance could be obtained through a cross-validation or increase the test dataset to reduce the bias. Other hyper-parameters, such as filters size, number of filters, generator-discriminator training scheme, and learning rate, can be further optimized through the evaluation strategies mentioned above.   

\section{Conclusion}
In this work, we developed and evaluated a deep network to generate virtual dual energy images from standard chest radiographs based on a multi-scale conditional adversarial network architecture (MCA-Net). With adversarial training, training of the multi-scale generator was reinforced so that it was able to learn how to preserve subtle signals in the virtual DE bone image. After obtaining the virtual bone image, high-quality virtual soft tissue images were produced by using a modified cross projection tensor algorithm to suppress the bone signal in the standard radiograph. Images from the testing data had high PSNR and SSIM values and were found to preserve clinically relevant features, indicating that the algorithm can produce virtual dual energy images comparable to ground truth dual energy images. The clinical evaluation demonstrated the clinical usefulness of this technique. Given the high cost of DES equipment and increased radiation dose from DE acquisitions, the use of virtual DE bone and soft tissue images provides a potential alternative solution for improved detection and diagnosis of disease using chest radiography.

\newpage
\bibliographystyle{splncs}
\bibliography{bibliography}

\begin{thebibliography}{10}

\bibitem{kelcz1994conventional}
Kelcz, F., Zink, F., Peppler, W., Kruger, D., Ergun, D., Mistretta, C.:
\newblock Conventional chest radiography vs dual-energy computed radiography in
  the detection and characterization of pulmonary nodules.
\newblock AJR. American journal of roentgenology \textbf{162} (1994)  271--278

\bibitem{li2011improved}
Li, F., Hara, T., Shiraishi, J., Engelmann, R., MacMahon, H., Doi, K.:
\newblock Improved detection of subtle lung nodules by use of chest radiographs
  with bone suppression imaging: receiver operating characteristic analysis
  with and without localization.
\newblock American Journal of Roentgenology \textbf{196} (2011)  W535--W541

\bibitem{zhou2016detection}
Zhou, B., Wen, D., Nye, K., Gilkeson, R.C., Eck, B., Jordan, D., Wilson, D.L.:
\newblock Detection and quantification of coronary calcium from dual energy
  chest x-rays: phantom feasibility study.
\newblock Medical physics (2016)

\bibitem{zhou2018visualization}
Zhou, B., Jiang, Y., Wen, D., Gilkeson, R.C., Hou, J., Wilson, D.L.:
\newblock Visualization of coronary artery calcium in dual energy chest
  radiography using automatic rib suppression.
\newblock In: Medical Imaging 2018: Image Processing. Volume 10574.,
  International Society for Optics and Photonics (2018)  105740E

\bibitem{wen2018enhanced}
Wen, D., Nye, K., Zhou, B., Gilkeson, R.C., Gupta, A., Ranim, S., Couturier,
  S., Wilson, D.L.:
\newblock Enhanced coronary calcium visualization and detection from dual
  energy chest x-rays with sliding organ registration.
\newblock Computerized Medical Imaging and Graphics (2018)

\bibitem{chen2013computerized}
Chen, S., Suzuki, K.:
\newblock Computerized detection of lung nodules by means of “virtual
  dual-energy” radiography.
\newblock IEEE Transactions on Biomedical Engineering \textbf{60} (2013)
  369--378

\bibitem{vock2009dual}
Vock, P., Szucs-Farkas, Z.:
\newblock Dual energy subtraction: principles and clinical applications.
\newblock European journal of radiology \textbf{72} (2009)  231--237

\bibitem{wang2017chestx}
Wang, X., Peng, Y., Lu, L., Lu, Z., Bagheri, M., Summers, R.M.:
\newblock Chestx-ray8: Hospital-scale chest x-ray database and benchmarks on
  weakly-supervised classification and localization of common thorax diseases.
\newblock In: 2017 IEEE Conference on Computer Vision and Pattern Recognition
  (CVPR), IEEE (2017)  3462--3471

\bibitem{shin2016deep}
Shin, H.C., Roth, H.R., Gao, M., Lu, L., Xu, Z., Nogues, I., Yao, J., Mollura,
  D., Summers, R.M.:
\newblock Deep convolutional neural networks for computer-aided detection: Cnn
  architectures, dataset characteristics and transfer learning.
\newblock IEEE transactions on medical imaging \textbf{35} (2016)  1285--1298

\bibitem{roth2015deeporgan}
Roth, H.R., Lu, L., Farag, A., Shin, H.C., Liu, J., Turkbey, E.B., Summers,
  R.M.:
\newblock Deeporgan: Multi-level deep convolutional networks for automated
  pancreas segmentation.
\newblock In: International Conference on Medical Image Computing and
  Computer-Assisted Intervention, Springer (2015)  556--564

\bibitem{isola2017image}
Isola, P., Zhu, J.Y., Zhou, T., Efros, A.A.:
\newblock Image-to-image translation with conditional adversarial networks.
\newblock arXiv preprint (2017)

\bibitem{agrawal2006edge}
Agrawal, A., Raskar, R., Chellappa, R.:
\newblock Edge suppression by gradient field transformation using
  cross-projection tensors.
\newblock In: Computer Vision and Pattern Recognition, 2006 IEEE Computer
  Society Conference on. Volume~2., IEEE (2006)  2301--2308

\bibitem{suzuki2004suppression}
Suzuki, K., Abe, H., Li, F., Doi, K.:
\newblock Suppression of the contrast of ribs in chest radiographs by means of
  massive training artificial neural network.
\newblock In: Medical Imaging 2004: Image Processing. Volume 5370.,
  International Society for Optics and Photonics (2004)  1109--1120

\bibitem{suzuki2006image}
Suzuki, K., Abe, H., MacMahon, H., Doi, K.:
\newblock Image-processing technique for suppressing ribs in chest radiographs
  by means of massive training artificial neural network (mtann).
\newblock IEEE Transactions on medical imaging \textbf{25} (2006)  406--416

\bibitem{chen2014separation}
Chen, S., Suzuki, K.:
\newblock Separation of bones from chest radiographs by means of anatomically
  specific multiple massive-training anns combined with total variation
  minimization smoothing.
\newblock IEEE transactions on medical imaging \textbf{33} (2014)  246--257

\bibitem{chen2016enhancement}
Chen, S., Zhong, S., Yao, L., Shang, Y., Suzuki, K.:
\newblock Enhancement of chest radiographs obtained in the intensive care unit
  through bone suppression and consistent processing.
\newblock Physics in Medicine \& Biology \textbf{61} (2016)  2283

\bibitem{loog2006filter}
Loog, M., van Ginneken, B., Schilham, A.M.:
\newblock Filter learning: application to suppression of bony structures from
  chest radiographs.
\newblock Medical image analysis \textbf{10} (2006)  826--840

\bibitem{simko2009elimination}
Simk{\'o}, G., Orb{\'a}n, G., M{\'a}day, P., Horv{\'a}th, G.:
\newblock Elimination of clavicle shadows to help automatic lung nodule
  detection on chest radiographs.
\newblock In: 4th European Conference of the International Federation for
  Medical and Biological Engineering, Springer (2009)  488--491

\bibitem{hogeweg2013suppression}
Hogeweg, L., Sanchez, C.I., van Ginneken, B.:
\newblock Suppression of translucent elongated structures: applications in
  chest radiography.
\newblock IEEE transactions on medical imaging \textbf{32} (2013)  2099--2113

\bibitem{rasheed2007rib}
Rasheed, T., Ahmed, B., Khan, M.A., Bettayeb, M., Lee, S., Kim, T.S.:
\newblock Rib suppression in frontal chest radiographs: A blind source
  separation approach.
\newblock In: Signal Processing and Its Applications, 2007. ISSPA 2007. 9th
  International Symposium on, IEEE (2007)  1--4

\bibitem{lecun2015deep}
LeCun, Y., Bengio, Y., Hinton, G.:
\newblock Deep learning.
\newblock nature \textbf{521} (2015)  436

\bibitem{mirza2014conditional}
Mirza, M., Osindero, S.:
\newblock Conditional generative adversarial nets.
\newblock arXiv preprint arXiv:1411.1784 (2014)

\bibitem{ledig2016photo}
Ledig, C., Theis, L., Husz{\'a}r, F., Caballero, J., Cunningham, A., Acosta,
  A., Aitken, A., Tejani, A., Totz, J., Wang, Z.,  et~al.:
\newblock Photo-realistic single image super-resolution using a generative
  adversarial network.
\newblock arXiv preprint (2016)

\bibitem{karras2017progressive}
Karras, T., Aila, T., Laine, S., Lehtinen, J.:
\newblock Progressive growing of gans for improved quality.
\newblock Stability, and Variation. arXiv preprint (2017)

\bibitem{ronneberger2015u}
Ronneberger, O., Fischer, P., Brox, T.:
\newblock U-net: Convolutional networks for biomedical image segmentation.
\newblock In: International Conference on Medical Image Computing and
  Computer-Assisted Intervention, Springer (2015)  234--241

\bibitem{goodfellow2014generative}
Goodfellow, I., Pouget-Abadie, J., Mirza, M., Xu, B., Warde-Farley, D., Ozair,
  S., Courville, A., Bengio, Y.:
\newblock Generative adversarial nets.
\newblock In: Advances in neural information processing systems. (2014)
  2672--2680

\bibitem{schilham2006computer}
Schilham, A.M., Van~Ginneken, B., Loog, M.:
\newblock A computer-aided diagnosis system for detection of lung nodules in
  chest radiographs with an evaluation on a public database.
\newblock Medical Image Analysis \textbf{10} (2006)  247--258

\bibitem{chakraborty2007spatial}
Chakraborty, D., Yoon, H.J., Mello-Thoms, C.:
\newblock Spatial localization accuracy of radiologists in free-response
  studies: inferring perceptual froc curves from mark-rating data.
\newblock Academic radiology \textbf{14} (2007)  4--18

\end{thebibliography}

\end{document}